\documentclass[10pt,twocolumn]{article}

\usepackage[letterpaper,margin=1in,columnsep=0.25in]{geometry}
\usepackage[T1]{fontenc}
\usepackage{amsmath}
\usepackage{amssymb}
\usepackage{amsfonts}
\usepackage{newtxtext}
\usepackage{newtxmath}
\usepackage{graphicx}
\usepackage[hyphens]{url}
\usepackage{booktabs}
\usepackage{multirow}
\usepackage{array}
\usepackage{enumitem}
\usepackage{nicefrac}
\usepackage{xcolor}
\usepackage{colortbl}
\usepackage{caption}
\usepackage{subcaption}
\usepackage{algorithm}
\usepackage{algorithmic}
\usepackage{authblk}
\usepackage[round]{natbib}
\usepackage[hidelinks]{hyperref}

\newcommand{\best}[1]{\textcolor{red}{#1}}
\newcommand{\sbest}[1]{\textcolor{blue}{\underline{#1}}}
\definecolor{lightpurple}{RGB}{230,225,245}

\title{\bfseries ThreshGuide: Class-Aware Labeled-Guided Thresholding for
Semi-Supervised 3D Abdominal Multi-Organ Segmentation}

\author[1]{Hongyu Liu}
\author[1]{Yinlong Wang}
\author[1]{Lusha Li}
\author[1]{Hui Meng$^{\ast}$}
\affil[1]{School of Intelligent Science and Technology, Hangzhou Institute
for Advanced Study, University of Chinese Academy of Sciences, Hangzhou, China}
\affil{\small liuhongyu24@mails.ucas.ac.cn \quad wangyinlng25@mails.ucas.ac.cn
\quad lilusha24@mails.ucas.ac.cn}
\affil{\small $^{\ast}$Corresponding author: huimeng@ucas.ac.cn}
\date{}

\begin{document}
\maketitle

\begin{abstract}
Pseudo-labeling is a strong paradigm for semi-supervised medical image segmentation, yet its effectiveness is highly sensitive to confidence thresholding.
In abdominal multi-organ segmentation, a fixed global threshold is particularly suboptimal because organ classes differ substantially in size, appearance, and learning difficulty.
In this work, we propose \textbf{ThreshGuide}, a class-aware threshold adaptation framework that uses labeled data to guide pseudo-label selection on unlabeled data.
Built upon a standard teacher-student architecture, the teacher model evaluates labeled samples during training to estimate class-aware threshold targets by maximizing an error-aware $F_{\beta}$ criterion that balances precision and coverage.
These targets are then smoothed with an exponential moving average (EMA) and used to filter unlabeled voxels in a class-dependent manner.
Experiments on FLARE2022 and AMOS2022 show that ThreshGuide performs competitively overall, yielding clear improvements specifically on hard-to-learn organs.
\end{abstract}

\section{Introduction}
Accurate multi-organ segmentation of abdominal Computed Tomography (CT) scans is a prerequisite for downstream quantitative analysis and clinical decision support.
However, acquiring dense pixel-level annotations is prohibitively expensive and relies heavily on domain expertise.
Consequently, Semi-Supervised Semantic Segmentation (SSSS) has emerged as a crucial technique to mitigate this annotation bottleneck by exploiting the information contained within large pools of unlabeled images.

One of the most important paradigms driving this progress is consistency regularization.
A highly representative method of this paradigm is FixMatch~\citep{sohn2020fixmatch}, which enforces cross-view consistency by generating pseudo-labels from weakly augmented unlabeled images and aligning the predictions of their strongly augmented counterparts.
Instead of utilizing all pseudo-labels blindly, such frameworks introduce high-confidence filtering to discard uncertain predictions.
By leveraging such pseudo label supervision, this paradigm facilitates effective learning from unlabeled data.

\begin{figure}
  \centering
  \includegraphics[width=0.48\textwidth]{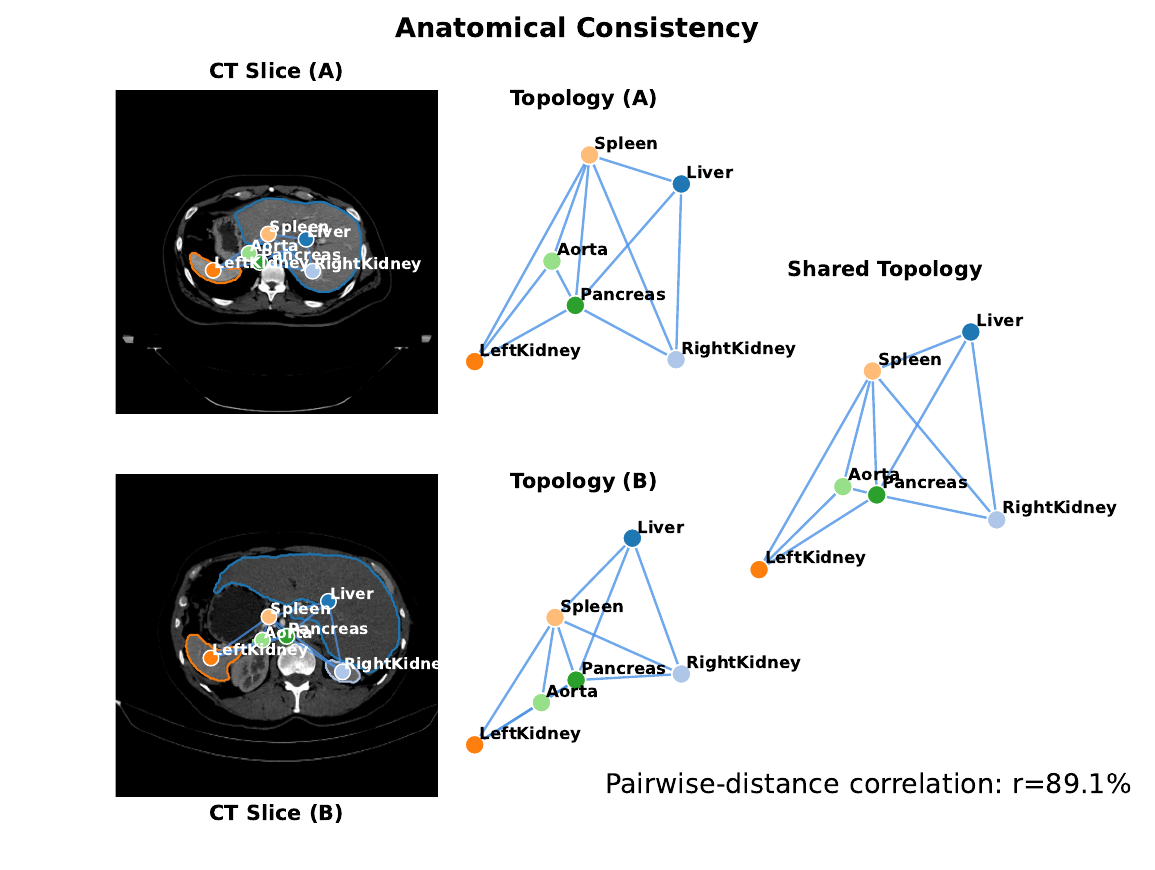}
  \caption{Topological consistency between labeled and unlabeled data.
  The pairwise organ distance correlation ($r=89.1\%$) after normalization reflects population-level topological consistency, suggesting highly aligned manifold semantic spaces that motivate using labeled data as a reliability proxy.}
  \label{fig:anatomical_consistency}
\end{figure}

Nevertheless, while a fixed high-confidence threshold helps ensure pseudo-label precision,
it compromises data utilization efficiency in tasks with strong inherent priors by discarding informative but low confidence predictions.
In abdominal multi-organ segmentation, a fixed global threshold induces a ``Matthew effect'': easy or dominant organs rapidly accumulate pseudo-label supervision, whereas difficult or underrepresented organs receive progressively less.
This homogeneous filtering largely overlooks the varying learning difficulties across different semantic classes.
As illustrated in Figure~\ref{fig:anatomical_consistency}, existing methods overlook a critical domain-specific prior~\citep{he2023geometric}: the high anatomical and distributional consistency of multi-organ structures between labeled and unlabeled abdominal scans.
In essence, this prior captures a more abstract and persistent regularity in how organs are spatially organized and related to one another, rooted in the shared biological blueprint of the human body.

\begin{figure}
  \centering
  \includegraphics[width=0.48\textwidth]{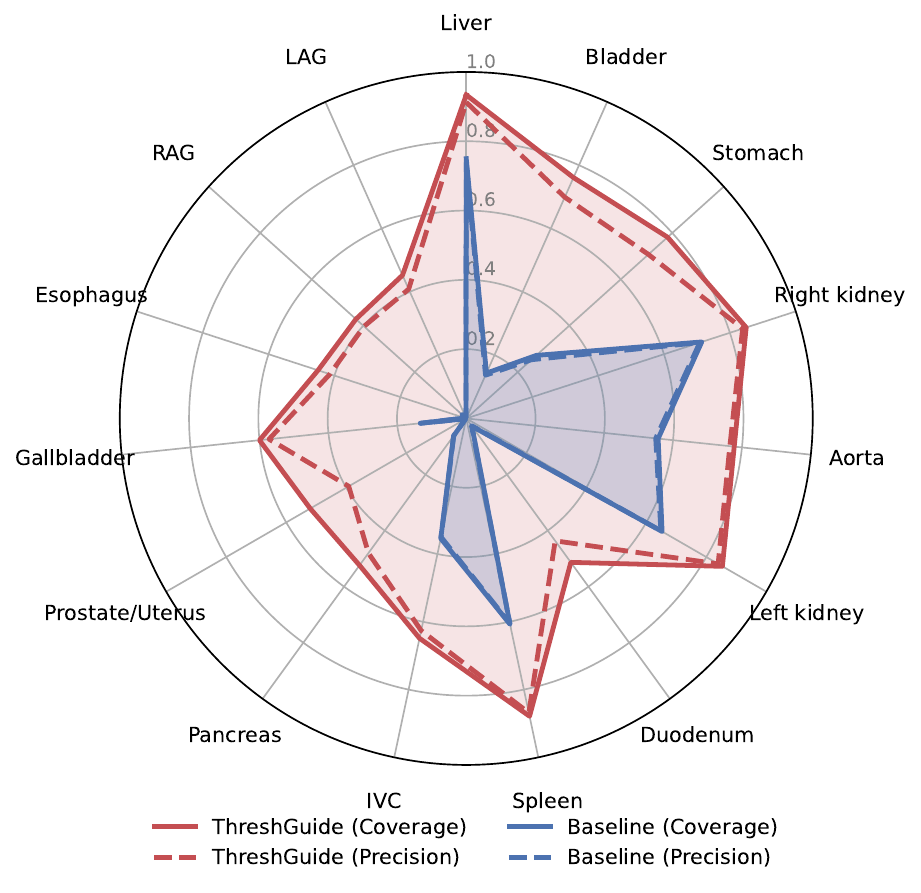}
  \caption{Coverage and Precision Radar. Organ-wise radar plot showing coverage (solid lines) and precision (dashed lines) for both the baseline and the proposed ThreshGuide. Because real unlabeled data do not have ground-truth annotations, the metrics are evaluated on the validation set as pseudo-unlabeled data.}
  \label{fig:motivation}
\end{figure}

Such fixed high confidence filtering is well suited to natural images, where objects exhibit substantial variability and complex priors in shape, texture, and color intensity.
However, in single-channel imaging data characterized by strong anatomical consistency
, such as abdominal multi-organ CT scans, inter-class confusion is typically less severe than in natural images
, while features derived from labeled and unlabeled samples tend to exhibit stronger correlations.
This discrepancy naturally prompts the question:
\textit{How can we exploit anatomical consistency to leverage labeled data, explicitly guiding the learning of unlabeled data?}

However, the fixed confidence threshold limits the supervisory coverage of unlabeled data
, as Figure~\ref{fig:motivation} clearly shows that only a small fraction of pixels in most classes receive supervision.
Motivated by this observation, we hypothesize that shared anatomical structures can be exploited to broaden supervisory coverage without compromising supervision reliability.
To operationalize this hypothesis, we leverage reliability signals derived from labeled data to guide pseudo-label selection on unlabeled data, leading to our proposed method, ThreshGuide.
Specifically, as shown in Figure~\ref{fig:motivation}
, by leveraging the optimal confidence threshold derived from labeled data as a labeled proxy
, the proposed strategy can explicitly guide pseudo-label selection for unlabeled data
, thereby achieving substantially higher precision and coverage compared with the conventional fixed-threshold scheme.

Moreover, subsequent experiments demonstrate that explicitly leveraging anatomical consistency outperforms methods that exploit low-confidence regions, such SoftMatch~\citep{chen2023softmatch}.
To alleviate the Matthew effect, we utilize the $F_\beta$ score to identify the optimal confidence threshold for each category based on labeled data
, thereby effectively balancing the precision and coverage of different categories.
Furthermore, to prevent dominant categories with excessive erroneous pseudo-labels from negatively affecting the learning of other categories
, we introduce an error-aware penalty mechanism to adaptively adjust the $\beta$ value in the $F_\beta$ formulation, enabling more efficient utilization of unlabeled data.
As illustrated in Figure~\ref{fig:cross_split_agreement}, this regularity also manifests in the confidence dynamics of semi-supervised learning: the labeled and unlabeled confidence trajectories are highly similar, sharing a latent common trend despite a clear distribution shift, while the unlabeled ones exhibit a noticeable lag behind the labeled ones.
Therefore, we further design an EMA update mechanism to alleviate such potential inconsistency and improve the stability of unlabeled data utilization.

In summary, this work makes three main contributions:
\begin{itemize}[
    leftmargin=1.35em,
    labelsep=0.45em,
    itemsep=1pt,
    topsep=2pt,
    parsep=0pt,
    partopsep=0pt
]
    \item We propose \textbf{ThreshGuide}, a semi-supervised segmentation
    framework that uses labeled data to guide class-aware pseudo-label
    filtering. The method replaces a fixed global threshold with dynamic
    class-aware thresholds, which is particularly suitable for highly
    imbalanced abdominal organ segmentation.

    \item We introduce \textbf{labeled-proxy calibration} based on teacher
    predictions on labeled data. The resulting threshold targets are obtained
    with a class-aware $F_{\beta}$ objective with error-aware weighting and
    stabilized through threshold EMA, yielding a simple and effective
    threshold estimation procedure.

    \item Extensive experiments on FLARE2022 and AMOS2022 demonstrate that
    ThreshGuide achieves state-of-the-art performance and consistently
    improves the segmentation accuracy of small organs.
\end{itemize}

\section{Related Work}
\subsection{Semi-Supervised Medical Image Segmentation}
Semi-supervised medical image segmentation mainly relies on consistency regularization and pseudo-labeling~\citep{jiang2022uncertainty,sohn2020fixmatch}.
Consistency regularization encourages the model to produce stable predictions under input perturbations or different augmentations, while pseudo-labeling turns confident predictions on unlabeled images into additional supervision.
The strongest results typically combine both in a weak-to-strong manner, where predictions from weakly-augmented views serve as pseudo-label to supervise strongly-augmented ones and thus improve robustness on unlabeled data.
Recent work further strengthens this paradigm through stronger augmentation, dynamic consistency control, and task-specific objectives.
DyCON~\citep{assefa2025dycon} dynamically adjusts the consistency weight during training to better balance early noisy supervision and later reliable regularization, whereas GALoss~\citep{qi2024gradient} is designed for abdominal multi-organ segmentation and alleviates severe class imbalance by combining class weighting and hard-sample mining within Dice and Cross-Entropy (CE) losses~\citep{shannon1948}.

\begin{figure}[t]
    \centering
    \makebox[\linewidth][c]{%
        \scalebox{1.15}[1.0]{%
            \includegraphics[
                width=0.7\linewidth
            ]{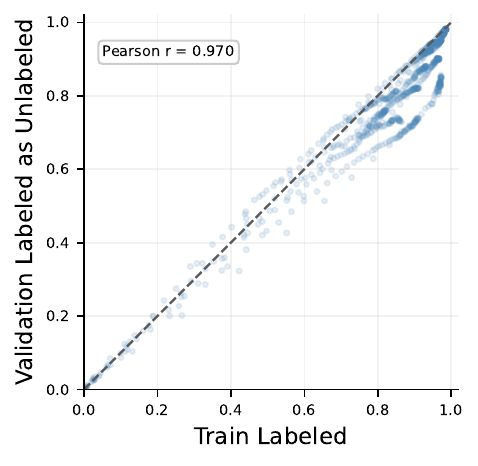}%
        }%
    }%
    \caption{
        Each point is one class mean confidence at one epoch.
        The Pearson correlation confirms consistent confidence
        , while points falling below the diagonal reflect the marginally higher teacher confidence on labeled data.}
    \label{fig:cross_split_agreement}
\end{figure}

\subsection{Adaptive Thresholding in Pseudo-Labeling}
Confidence thresholding is a widely used strategy in pseudo-labeling to filter unlabeled predictions based on their confidence scores, thereby controlling pseudo-label quality.
FixMatch~\citep{sohn2020fixmatch} uses a fixed threshold, which often underutilizes unlabeled data and exacerbates pseudo-label imbalance in abdominal multi-organ segmentation.
Adaptive variants address this issue by estimating thresholds from unlabeled predictions.
FreeMatch~\citep{wang2023freematch} introduces a self-adaptive thresholding mechanism that adjusts pseudo-label selection according to the model's learning status, with class-aware modulation to alleviate class imbalance.
SoftMatch~\citep{chen2023softmatch} addresses the quantity-quality trade-off by modeling pseudo-label confidence with soft weighting rather than a rigid hard cutoff.
However, both methods estimate their selection criteria mainly from unlabeled prediction statistics and do not explicitly use labeled data to calibrate threshold evolution.
For abdominal multi-organ segmentation, where labeled and unlabeled scans share strong anatomical structure, this leaves potentially useful supervisory information underexploited.
Our method addresses this gap by using labeled data to guide class-aware threshold evolution on unlabeled data in probability space.

\section{Methodology}
\label{sec:method}

\begin{figure*}[t]
  \centering
  \includegraphics[width=0.92\textwidth]{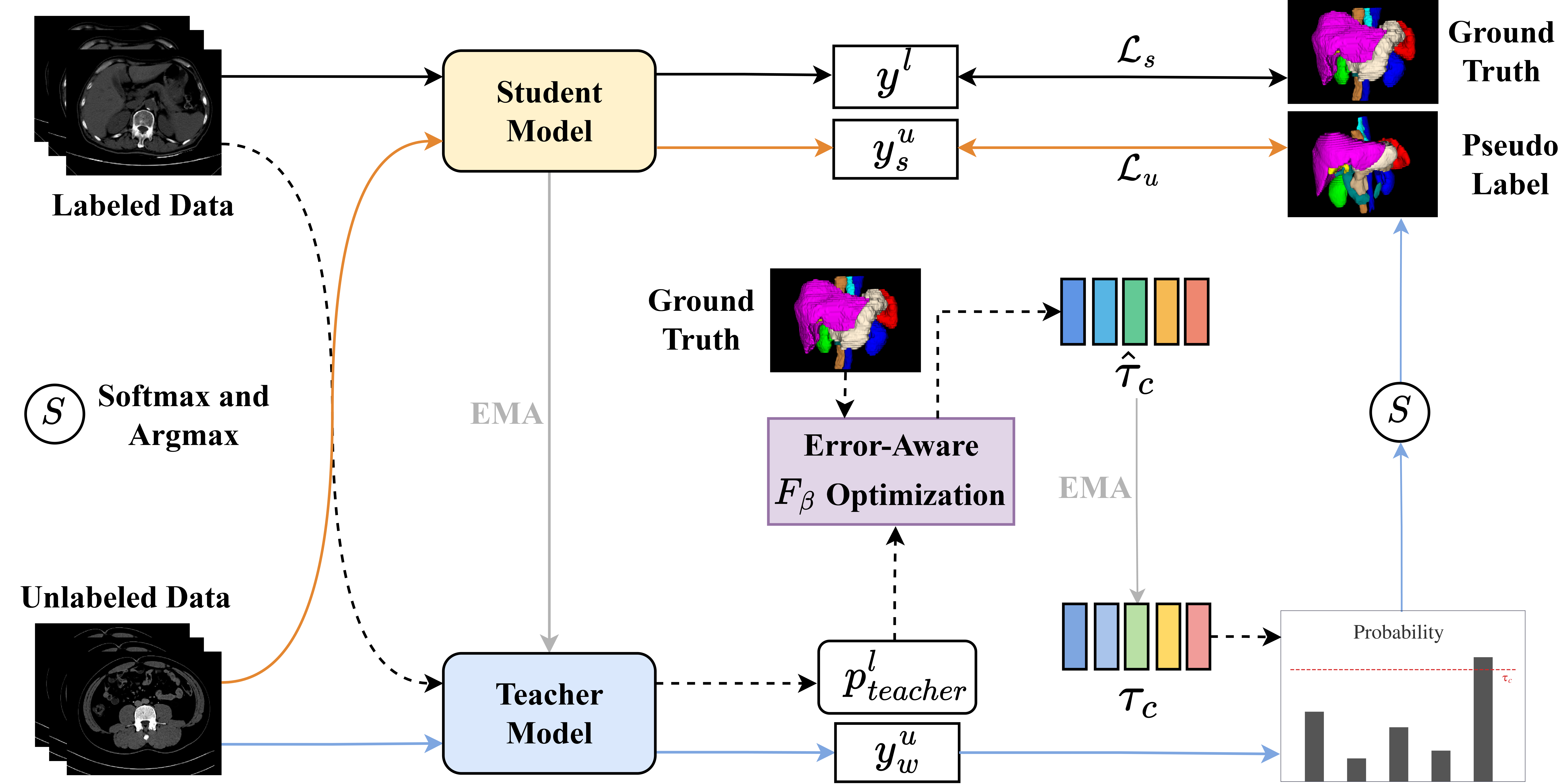}
  \caption{\textbf{Overview of ThreshGuide.}
  The teacher evaluates labeled samples without gradient propagation and
  exposes a class-wise confidence--correctness relationship. A
  precision-coverage criterion produces labeled-proxy threshold targets,
  while class-dependent $\beta_c$ weighting controls the preference between
  reliable precision and usable coverage. The targets are temporally smoothed
  and applied to the teacher predictions of weakly augmented unlabeled data,
  which supervise the student predictions of strongly augmented views.}
  \label{fig:framework}
\end{figure*}
Pseudo-label supervision provides a powerful mechanism of exploiting unlabeled data.
High-confidence filtering generally provides effective supervision when leveraging unlabeled data.
 However, Figure~\ref{fig:cross_split_agreement} demonstrates that, in abdominal multi-organ segmentation
 , labeled and unlabeled data exhibit highly consistent class-wise confidence trajectories throughout training.
This cross-split consistency motivates ThreshGuide's \emph{labeled-proxy calibration}
, which uses observable labeled data to calibrate pseudo-label selection on unlabeled data.
 At each training iteration, the
teacher predicts the labeled samples, where both prediction confidence
and correctness are observable.
Based on these observations, we
estimate for each class a proxy threshold that balances the precision
and coverage of the selected predictions.
Given the shared anatomical
structure and task distribution between labeled and unlabeled
multi-organ scans, the thresholds estimated from labeled data provide
a reliability reference for pseudo-label selection on unlabeled data.
We further update the class-wise thresholds through EMA, which reduces the variance of mini-batch estimates and
produces a stable threshold evolution throughout training. The overall
framework of ThreshGuide is illustrated in Figure~\ref{fig:framework}.

\subsection{Problem Setup and Teacher--Student Baseline}

Let $\mathcal{D}^{l}=\{(x_i^{l},y_i^{l})\}_{i=1}^{N^{l}}$ and
$\mathcal{D}^{u}=\{x_i^{u}\}_{i=1}^{N^{u}}$ denote the labeled and unlabeled
sets, respectively, where $N^{u}\gg N^{l}$ and the segmentation task contains
$C$ classes. We adopt a student network $f_{\theta}$ and an EMA teacher
$f_{\phi}$. The teacher parameters follow the student throughout training:
\begin{equation}
    \phi_t = \alpha_t \phi_{t-1} + (1-\alpha_t)\theta_t,
    \label{eq:teacher_ema}
\end{equation}
where $\alpha_t$ is the teacher momentum. For a labeled sample, the supervised
objective combines cross-entropy and Dice losses:
\begin{equation}
    \mathcal{L}_{s}
    = \frac{1}{2}\left[
    \mathcal{L}_{CE}(f_{\theta}(x^{l}),y^{l})
    +\mathcal{L}_{Dice}(f_{\theta}(x^{l}),y^{l})
    \right].
    \label{eq:sup_loss}
\end{equation}

For an unlabeled sample $x^{u}$, the teacher predicts the weakly augmented view
and the student predicts the strongly augmented view:
$p_w^{u}=\operatorname{softmax}(f_{\phi}(\mathcal{A}_w(x^{u})))$ and
$p_s^{u}=\operatorname{softmax}(f_{\theta}(\mathcal{A}_s(x^{u})))$.
The teacher pseudo-label is
$\hat y_w^{u(v)}=\arg\max_c p_{w,c}^{u(v)}$. Unlike FixMatch, which applies one
predefined cutoff to every class, ThreshGuide calibrates the acceptance
threshold from labeled teacher behavior.

\subsection{Labeled-Proxy Calibration for Pseudo-Label}
\label{sec:dynamic_thresholding}

\paragraph{Balancing precision and coverage.}
Given that the consistent confidence evolution observed in
Figure~\ref{fig:cross_split_agreement}
 indirectly suggests anatomical correspondence between labeled and unlabeled data.
We consider a labeled voxel $v$, let $p_T^{l(v)}$ denote the teacher probability vector,
$\hat y_T^{l(v)}$ its predicted class, and $q_T^{l(v)}$ its maximum confidence.
Because $y^{l(v)}$ is known, the correctness associated with each confidence is
directly observable. For class $c$, we collect the voxels predicted as $c$ and
sort their confidence values in descending order,
$q_{c,(1)}\geq\cdots\geq q_{c,(N_c)}$. Considering the top-$k$ predictions as a
candidate accepted set, we define
\begin{equation}
\operatorname{Pre}_{c}(k)=\frac{TP_c(k)}{k},
\qquad
\operatorname{Cov}_{c}(k)=\frac{TP_c(k)}{N_c},
\label{eq:pre_cov}
\end{equation}
where $TP_c(k)$ is the number of correct predictions among the top-$k$ elements
and $N_c$ is the size of the teacher-predicted class-$c$ pool. Here,
$\operatorname{Cov}_{c}$ measures how much verified class-$c$ evidence is
retained from that pool; it is used together with precision to avoid selecting
either a tiny but overly conservative set or a large but noisy set.

We score each candidate through a class-dependent precision-coverage utility:
\begin{equation}
F_{\beta_c}(k)=
\frac{(1+\beta_c^2)\operatorname{Pre}_{c}(k)
\operatorname{Cov}_{c}(k)}
{\beta_c^2\operatorname{Pre}_{c}(k)+\operatorname{Cov}_{c}(k)}.
\label{eq:fbeta_proxy}
\end{equation}
The labeled-proxy target is determined by
\begin{equation}
    k_c^{*}=\arg\max_k F_{\beta_c}(k),
    \qquad
    \hat\tau_c^{(t)}=q_{c,(k_c^{*})}.
\label{eq:proxy_target}
\end{equation}
Thus, the threshold is not manually scheduled: it follows the operating point
at which the current teacher achieves the most suitable balance between
correctness and usable supervision for class $c$.

\begin{figure}[t]
      \centering
      \includegraphics[
          width=0.92\linewidth
      ]{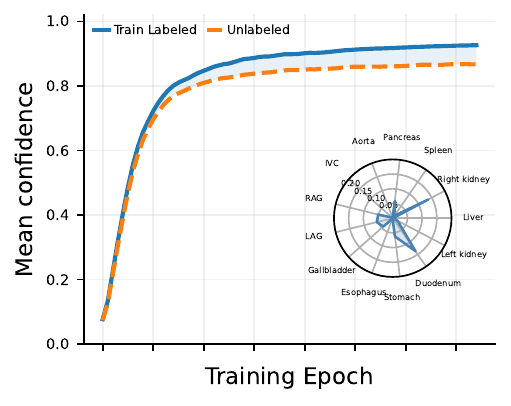}
      \caption{
      Class-wise mean confidence trajectories on the labeled training set and
      the validation set used as a proxy for unlabeled data, together with their
      corresponding per-class confidence gaps.
      }
      \label{fig:class_confidence_trajectory}
  \end{figure}

\paragraph{Class-dependent preference for precision.}
As shown in Figure~\ref{fig:class_confidence_trajectory}
, these organ-dependent confidence dynamics indicate that the same precision-coverage preference is not appropriate for all abdominal structures.
If such errors are admitted without control, they can occupy a large part of the pseudo-supervision.
ThreshGuide therefore uses $\beta_c$ to shift the proxy
search toward precision whenever the observed class-wise error burden is high.

Let $r_c^{(t)}$ be the proportion of teacher-predicted foreground voxels assigned
to class $c$, and let $\mu_c^{(t)}$ be its EMA estimate:
\begin{equation}
 r_c^{(t)}=
 \frac{\sum_{v\in\Omega_l}\mathbb{I}(\hat y_T^{l(v)}=c)}
 {\sum_{v\in\Omega_l}\mathbb{I}(\hat y_T^{l(v)}>0)},
 \,
 \mu_c^{(t)}=\rho_t\mu_c^{(t-1)}+(1-\rho_t)r_c^{(t)}.
\label{eq:class_occupancy}
\end{equation}
We further measure the normalized teacher error in the predicted class pool,
$\delta_c^{(t)}=E_c^{(t)}/N_c$, where $E_c^{(t)}$ counts its incorrect
predictions. The class-dependent weight is
\begin{equation}
    \beta_c^{(t)}=
    \frac{1}{1-\log(\mu_c^{(t)})}
    \exp\!\left(-\delta_c^{(t)}\right).
\label{eq:beta_final}
\end{equation}
The occupancy term adapts the utility to heterogeneous class scales, whereas
the error term decreases $\beta_c$ when unreliable predictions accumulate.
Since a smaller $\beta_c$ makes Eq.~\eqref{eq:fbeta_proxy} more
precision-oriented, the selected threshold becomes more conservative when
false pseudo-labels would otherwise dominate the accepted set.
A formal analysis of the mathematical properties of each component in Eq.~\eqref{eq:beta_final}
, together with a minimax interpretation of the overall objective, is provided in Appendix~\ref{sec:appendix_B}.

\paragraph{EMA-smoothed threshold for unlabeled data.}
Figure~\ref{fig:class_confidence_trajectory} illustrates that the class-wise mean confidence on the labeled training set increases faster than that on the validation set treated as proxy-unlabeled data
, while the magnitude and temporal evolution of the resulting cross-split confidence gap vary substantially across organs.
This gap is expected because the labeled samples directly participate in model optimization
, whereas the held-out validation samples are not used for gradient-based parameter updates; consequently
, their confidence rises more slowly and remains lower during the later stages of training.
Accordingly, we adopt an EMA scheme to update the class-wise threshold using the proxy target $\hat{\tau}_c^{(t)}$
, and we regard the target as
\begin{equation}
    \tau_c^{(t)}=
    \alpha_{\tau}\tau_c^{(t-1)}
    +(1-\alpha_{\tau})\hat\tau_c^{(t)}.
\label{eq:threshold_ema}
\end{equation}
This temporal transfer suppresses mini-batch fluctuations, preserves the
long-term class-wise trend, and avoids forcing the richer unlabeled data to
follow every short-lived change observed on labeled samples. If class $c$ is
absent from the current labeled batch, its previous threshold is retained.

\subsection{Proxy-Guided Consistency Learning}
\label{sec:unsupervised_consistency}

For an unlabeled voxel $v$, the threshold associated with its teacher-predicted
class is used to determine whether it provides supervision:
\begin{equation}
    \mathcal{M}^{(v)}=
    \mathbb{I}\!\left(
    \max_c p_{w,c}^{u(v)}\geq
    \tau_{\hat y_w^{u(v)}}^{(t)}
    \right).
\label{eq:class_mask}
\end{equation}
The weak-view pseudo-label then supervises the strongly augmented prediction
only at accepted voxels:
\begin{equation}
    \mathcal{L}_{u}
    =\frac{1}{|\Omega_u|}
    \sum_{v\in\Omega_u}
    \mathcal{M}^{(v)}
    \mathcal{L}_{CE}\!\left(
    p_s^{u(v)},\hat y_w^{u(v)}
    \right).
\label{eq:unsup_loss}
\end{equation}
The complete training objective is
\begin{equation}
    \mathcal{L}_{total}=\mathcal{L}_{s}+\lambda_u\mathcal{L}_{u},
\label{eq:total_loss}
\end{equation}
where $\lambda_u$ controls the contribution of unlabeled consistency.
The complete training procedure is summarized in Algorithm~\ref{alg:threshguide} of Appendix~\ref{sec:appendix_A}.

\section{Experiments}
\label{sec:experiments}

\begin{table*}
    \centering
    \setlength{\tabcolsep}{1.8pt}
    \resizebox{\textwidth}{!}{
    \begin{tabular}{l *{15}{r} >{\columncolor{lightpurple}}r >{\columncolor{lightpurple}}r}
        \toprule
        \rowcolor{green!6}
        & \multicolumn{15}{c}{\textbf{Dice Score for Each Organ}} & \multicolumn{2}{c}{\textbf{Average}} \\
        \cmidrule(lr){2-16} \cmidrule(lr){17-18}
        \rowcolor{green!6}
        \multirow{-3}{*}{\textbf{Methods}} & Spl & R.kid & L.kid & Gal & Eso & Liv & Sto & Aor & IVC & Pan & RAG & LAG & Duo & Bla & P/U & \textbf{Dice} & \textbf{Jac.} \\
        \midrule
        \multicolumn{18}{c}{\textbf{240 Labeled / 1200 Unlabeled (1:5 Ratio)}} \\
        \midrule
        SupOnly~\citep{cciccek20163d}      & \sbest{95.92}& 95.59        & 92.38        & 65.99        & 0.00         & \sbest{97.32}& 89.34        & 93.91        & 86.60        & 78.32        & 0.00         & 0.00         & 71.25        & 81.71        & 73.28        & 68.11 $\pm$ 0.29        & 60.84 $\pm$ 0.23 \\
        MT~\citep{tarvainen2017mean}       & 95.39        & 95.22        & 91.93        & 71.87        & 63.63        & 97.01        & 88.75        & 93.35        & 87.06        & 78.18        & 13.90        & 26.93        & 70.47        & 81.67        & 71.50        & 75.12 $\pm$ 1.23        & 66.44 $\pm$ 1.13 \\
        FixMatch~\citep{sohn2020fixmatch}  & 95.66        & 95.59        & 92.40        & 76.21        & 76.52        & 97.16        & 88.63        & 93.85        & 87.22        & 79.89        & 66.39        & \sbest{64.96}& 73.55        & 82.19        & 73.16        & 82.89 $\pm$ 0.18        & 74.29 $\pm$ 0.15 \\
        GA-MT~\citep{qi2024gradient}       & 94.25        & 93.43        & 90.38        & \best{81.17} & 72.24        & 96.78        & \sbest{89.49}& 90.86        & 85.80        & 79.78        & 58.57        & 58.68        & 73.27        & 80.79        & 70.74        & 81.08 $\pm$ 0.27        & 71.45 $\pm$ 0.26 \\
        CPS~\citep{chen2021semi}           & 95.32        & 95.10        & 91.90        & 71.75        & 62.39        & 96.98        & 88.19        & 93.43        & 87.03        & 78.15        & 33.49        & 39.14        & 69.94        & 81.34        & 70.88        & 77.00 $\pm$ 0.85        & 67.97 $\pm$ 0.60 \\
        DyCON~\citep{assefa2025dycon}      & 95.52        & 95.28        & 92.08        & 72.66        & 53.00        & 97.03        & 87.88        & 93.69        & 86.22        & 77.60        & 8.31         & 27.44        & 68.21        & \sbest{82.77}& 72.96        & 74.04 $\pm$ 0.76        & 65.22 $\pm$ 0.69 \\
        SoftMatch~\citep{chen2023softmatch}& 95.60        & \sbest{95.60}& \sbest{92.43}& 78.69        & 76.40        & 97.18        & 88.69        & 93.80        & 87.17        & 79.29        & \sbest{66.83}& 63.54        & 73.21        & 82.32        & 73.95        & 82.98 $\pm$ 0.34        & 74.31 $\pm$ 0.34 \\
        FreeMatch~\citep{wang2023freematch}& 95.73        & 95.54        & 92.38        & 78.40        & \sbest{76.88}& 97.20        & 88.90        & \sbest{93.92}& \sbest{87.35}& \sbest{79.92}& 66.67        & 64.84        & \sbest{73.85}& 82.56        & \sbest{74.46}& \sbest{83.24 $\pm$ 0.11}& \sbest{74.62 $\pm$ 0.18}\\
        Ours                              & \best{96.22} & \best{95.93} & \best{92.78} & \sbest{80.65}& \best{80.02} & \best{97.55} & \best{90.47} & \best{94.34} & \best{88.49} & \best{82.30} & \best{70.78} & \best{71.93} & \best{76.89} & \best{84.02} & \best{77.53} & \best{85.33 $\pm$ 0.07} & \best{77.26 $\pm$ 0.18} \\
        \midrule
        \multicolumn{18}{c}{\textbf{80 Labeled / 1200 Unlabeled (1:15 Ratio)}} \\
        \midrule
        SupOnly~\citep{cciccek20163d}      & \sbest{95.87}& 95.43        & 91.98        & 0.00         & 0.00         & 96.92        & \sbest{88.82}& 92.91        & 81.87        & 74.55        & 0.00         & 0.00         & 65.64        & 76.21        & 42.47        & 60.18 $\pm$ 2.28        & 55.16 $\pm$ 0.45 \\
        MT~\citep{tarvainen2017mean}       & 95.84        & 95.31        & 92.28        & 71.33        & 47.87        & 97.06        & 88.72        & 93.42        & 86.11        & 75.10        & 0.61         & 30.17        & 66.52        & 81.73        & 65.23        & 72.49 $\pm$ 1.23        & 64.18 $\pm$ 1.02 \\
        FixMatch~\citep{sohn2020fixmatch}  & 95.82        & 95.55        & 92.44        & 76.57        & 74.26        & 97.10        & 88.70        & \sbest{93.76}& 87.10        & 77.44        & 66.44        & 65.29        & 70.92        & \sbest{82.97}& 69.88        & 82.28 $\pm$ 0.39        & 73.72 $\pm$ 0.39 \\
        GA-MT~\citep{qi2024gradient}       & 93.96        & 93.20        & 90.25        & 77.46        & 68.34        & 96.58        & 88.57        & 88.76        & 82.96        & 76.36        & 55.64        & 52.44        & 68.72        & 80.75        & 68.82        & 78.85 $\pm$ 0.58        & 68.47 $\pm$ 0.70 \\
        CPS~\citep{chen2021semi}           & 95.72        & 95.33        & 92.23        & 67.68        & 42.00        & 96.94        & 88.08        & 92.58        & 85.50        & 75.16        & 5.91         & 31.11        & 65.59        & 79.78        & 59.64        & 71.55 $\pm$ 0.78        & 63.54 $\pm$ 0.67 \\
        DyCON~\citep{assefa2025dycon}      & 95.63        & 95.23        & 92.14        & 68.53        & 45.17        & 96.96        & 87.57        & 93.44        & 85.99        & 75.00        & 0.26         & 21.19        & 65.32        & 81.53        & 63.02        & 71.13 $\pm$ 0.25        & 62.88 $\pm$ 0.10 \\
        SoftMatch~\citep{chen2023softmatch}& 95.57        & 95.53        & \sbest{92.49}& 77.84        & \sbest{74.65}& 97.04        & 88.07        & 93.58        & 86.95        & 77.64        & \sbest{67.06}& 65.41        & 71.50        & 81.50        & 67.60        & 82.16 $\pm$ 0.29        & 73.78 $\pm$ 0.46 \\
        FreeMatch~\citep{wang2023freematch}& 95.78        & \sbest{95.56}& 92.45        & \sbest{78.61}& 74.20        & \best{97.12}& 87.98        & 93.73        & \sbest{87.11}& \sbest{78.21}& 66.20        & \sbest{65.69}& \best{73.20} & 82.78        & \sbest{71.11}& \sbest{82.65 $\pm$ 0.32}& \sbest{74.04 $\pm$ 0.43}\\
        Ours                              & \best{96.04} & \best{95.62} & \best{92.66} & \best{80.01} & \best{77.57} & \best{97.38} & \best{89.47} & \best{93.90} & \best{87.83} & \best{79.33} & \best{70.64} & \best{70.38} & \sbest{73.04}& \best{83.70} & \best{72.44} & \best{84.00 $\pm$ 0.14} & \best{75.71 $\pm$ 0.08} \\
        \bottomrule
    \end{tabular}
    }
    \caption{Quantitative comparison (class-wise Dice scores, and the overall Dice and Jaccard indices reported as mean $\pm$ standard deviation) on AMOS2022 datasets.
    Comparison of different methods under 1:5 and 1:15 labeled ratios. We report detailed scores for all 15 organs.
    Note: ``Jac.'' denotes Jaccard.
    The \best{best} and \sbest{second-best} results are in \best{red} and \sbest{underlined blue}, respectively.}
    \label{tab:main_results_amos}
\end{table*}

\subsection{Experimental Setup}

\textbf{Datasets.} We evaluate our method on two public abdominal multi-organ segmentation datasets.
(1) FLARE2022~\citep{ma2024unleashing} comprises 100 labeled and 2000 unlabeled CT scans covering 13 organ classes (with one background): the liver (Liv), spleen (Spl), pancreas (Pan), right kidney (R.kid), left kidney (L.kid), stomach (Sto), gallbladder (Gal), esophagus (Eso), aorta (Aor), inferior vena cava (IVC), right adrenal gland (RAG), left adrenal gland (LAG), and duodenum (Duo).
The labeled scans are split into 60, 20, and 20 for training, validation, and testing, respectively.
(2) AMOS2022~\citep{ji2022amos} is a 16-class segmentation dataset targeting 15 anatomical structures, including two additional organs not found in FLARE2022: the bladder (Bla) and prostate/uterus (P/U).
Its 300 labeled scans are partitioned into 240, 30, and 30 for training, validation, and testing, respectively. In addition, the dataset includes 1200 unlabeled scans.

\textbf{Data Preprocessing.}
All CT scans undergo a standardized preprocessing pipeline, which includes orientation standardization, HU windowing, intensity normalization, and spatial resampling.
Detailed preprocessing procedures are provided in Appendix~\ref{sec:appendix_A}.

\textbf{Implementation Details.}
To ensure a fair comparison, we adopt the standard 3D U-Net as the unified backbone architecture for all image segmentation experiments.
We implement our ThreshGuide framework using PyTorch~\citep{paszke2019pytorch} and conduct all experiments on a single NVIDIA RTX 4090 GPU.
The network is optimized using AdamW~\citep{loshchilov2017decoupled} with a polynomial learning rate schedule.
During training, we sample balanced labeled and unlabeled crops in each batch. In our experiments, this corresponds to a total batch size of $8$ (comprising $4$ labeled and $4$ unlabeled 3D volume crops).
The models are trained for a total of 37,500 and 45,000 iterations on the FLARE2022 and AMOS2022 datasets, respectively.
Additional implementation details, including hyperparameters, are provided in Appendix~\ref{sec:appendix_A}.

\subsection{Comparison with State-of-the-Arts}
\textbf{Compared Methods.}
We compare our proposed method with both established baselines and recent state-of-the-art semi-supervised segmentation approaches, including:
Mean Teacher (MT)~\citep{tarvainen2017mean}, FixMatch~\citep{sohn2020fixmatch}, GALoss with MT~\citep{qi2024gradient}, CPS~\citep{chen2021semi}, DyCON~\citep{assefa2025dycon}, SoftMatch~\citep{chen2023softmatch} and FreeMatch~\citep{wang2023freematch}.
For fairness, all methods are evaluated under the same data preprocessing, backbone, and dataset partition protocol, while the method-specific training settings follow the standard configurations reported in their original papers whenever applicable.

\begin{table*}[t]
    \centering
    \setlength{\tabcolsep}{4pt}
    \resizebox{\textwidth}{!}{
    \begin{tabular}{l >{\columncolor{lightpurple}}c >{\columncolor{lightpurple}}c cccccc >{\columncolor{lightpurple}}c >{\columncolor{lightpurple}}c cccccc}
        \toprule
        \rowcolor{green!6}
        & \multicolumn{8}{c}{\textbf{60:2000 (1:33 Ratio)}} & \multicolumn{8}{c}{\textbf{40:2000 (1:50 Ratio)}} \\
        \cmidrule(lr){2-9} \cmidrule(lr){10-17}
        \rowcolor{green!6}
        \multirow{-2}{*}{\textbf{Methods}} & Dice & Jac. & RAG & LAG & Gal & Eso & Sto & Duo & Dice & Jac. & RAG & LAG & Gal & Eso & Sto & Duo \\
        \midrule
        SupOnly~\citep{cciccek20163d}       & 68.75 $\pm$ 0.50        & 63.21 $\pm$ 0.96        & 0.00         & 0.00         & 78.49        & 0.00         & \best{87.32} & 74.63        & 67.00 $\pm$ 0.29        & 60.73 $\pm$ 0.58        & 0.00         & 0.00         & 79.11        & 0.00         & 83.54        & 70.86 \\
        MT~\citep{tarvainen2017mean}        & 85.51 $\pm$ 0.25        & 76.62 $\pm$ 0.13        & 69.62        & 69.25        & 86.58        & 69.70        & 86.03        & 77.27        & 83.07 $\pm$ 0.59        & 73.63 $\pm$ 0.50        & 65.59        & 68.00        & 81.88        & 71.37        & 81.80        & 74.66 \\
        FixMatch~\citep{sohn2020fixmatch}   & 85.22 $\pm$ 0.39        & 76.25 $\pm$ 0.39        & 67.30        & 70.31        & 86.68        & 69.88        & 86.64        & 75.95        & 83.61 $\pm$ 0.15        & 74.31 $\pm$ 0.22        & 61.87        & 67.22        & 83.04        & 73.28        & \sbest{85.37}& 74.10 \\
        GA-MT~\citep{qi2024gradient}        & 85.97 $\pm$ 0.34        & 76.43 $\pm$ 0.40        & 74.18        & 73.69        & 85.87        & 72.38        & 85.51        & 80.06        & 83.80 $\pm$ 0.75        & 73.58 $\pm$ 0.95        & 73.80        & 70.68        & 82.22        & 72.77        & 81.69        & 72.92 \\
        CPS~\citep{chen2021semi}            & 84.82 $\pm$ 0.39        & 76.17 $\pm$ 0.66        & 66.29        & 68.75        & 84.45        & 68.85        & 86.71        & 74.71        & 81.54 $\pm$ 0.06        & 72.19 $\pm$ 0.37        & 62.81        & 61.91        & 77.05        & 67.57        & 83.06        & 72.39 \\
        DyCON~\citep{assefa2025dycon}       & 83.34 $\pm$ 0.24        & 74.53 $\pm$ 0.33        & 61.50        & 58.04        & 85.57        & 65.74        & 86.61        & 73.88        & 81.73 $\pm$ 0.45        & 72.56 $\pm$ 0.22        & 61.55        & 59.38        & 81.28        & 68.69        & 84.35        & 70.64 \\
        SoftMatch~\citep{chen2023softmatch} & 88.07 $\pm$ 0.13        & 79.88 $\pm$ 0.35        & 78.35        & \sbest{79.55}& 88.95        & 75.53        & 86.93        & 79.24        & \sbest{86.11 $\pm$ 0.11}& \sbest{77.07 $\pm$ 0.13}& \sbest{76.81}& \sbest{74.57}& \sbest{86.66}& \best{76.55} & 83.57        & \sbest{75.43}\\
        FreeMatch~\citep{wang2023freematch} & \sbest{88.22 $\pm$ 0.17}& \sbest{79.90 $\pm$ 0.25}& \sbest{78.53}& 78.95        & \sbest{89.50}& \sbest{76.09}& 86.64        & \sbest{80.34}& 85.46 $\pm$ 0.23        & 76.31 $\pm$ 0.25        & 74.54        & 74.31        & 85.03        & 76.29        & 83.10        & 74.33 \\
        Ours                               & \best{88.79 $\pm$ 0.20} & \best{80.98 $\pm$ 0.28} & \best{79.87} & \best{80.59} & \best{89.70} & \best{78.11} & \sbest{87.18}& \best{80.94} & \best{87.18 $\pm$ 0.11} & \best{78.36 $\pm$ 0.19} & \best{76.95} & \best{77.25} & \best{89.06} & \sbest{76.49}& \best{85.45} & \best{78.30} \\
        \bottomrule
    \end{tabular}
    }
    \caption{Quantitative comparison (class-wise Dice scores, and the overall Dice and Jaccard indices reported as mean $\pm$ standard deviation) on FLARE2022 datasets.
    Comparison of different methods under 1:33 and 1:50 labeled ratios.
    The \best{best} and \sbest{second-best} results are highlighted in \best{red} and \sbest{underlined blue}, respectively.}
    \label{tab:main_results_flare}
\end{table*}

\textbf{Main Results.}
As demonstrated in Table~\ref{tab:main_results_amos}, our proposed ThreshGuide consistently achieves state-of-the-art performance on the AMOS2022 benchmark across all labeling ratios (1:5 and 1:15).
Under the 1:15 ratio, ThreshGuide attains the highest average Dice of 84.00\%, yielding a substantial +1.35\% improvement over FreeMatch.
Notably, ThreshGuide achieves Dice scores of 70.64\% for RAG and 70.38\% for LAG, surpassing FreeMatch by 4.44 and 4.69 percentage points, respectively.
The improvements on difficult classes such as LAG, RAG, and Duo validate the efficacy of our dynamically adjusted thresholding mechanism in handling class imbalance.

\textbf{Results on FLARE2022.}
To further validate the effectiveness of our method, we additionally evaluate on the FLARE2022 benchmark under semi-supervised settings with scarce labeled data (labeled ratios of 1:33 and 1:50).
As shown in Table~\ref{tab:main_results_flare}, ThreshGuide consistently achieves the best overall performance across both labeling ratios.
This superiority is maintained under the more extreme 1:50 ratio (40 labeled cases)
, confirming the effectiveness of our method in leveraging unlabeled data.

\begin{figure*}[t]
  \centering
  \includegraphics[width=0.98\textwidth]{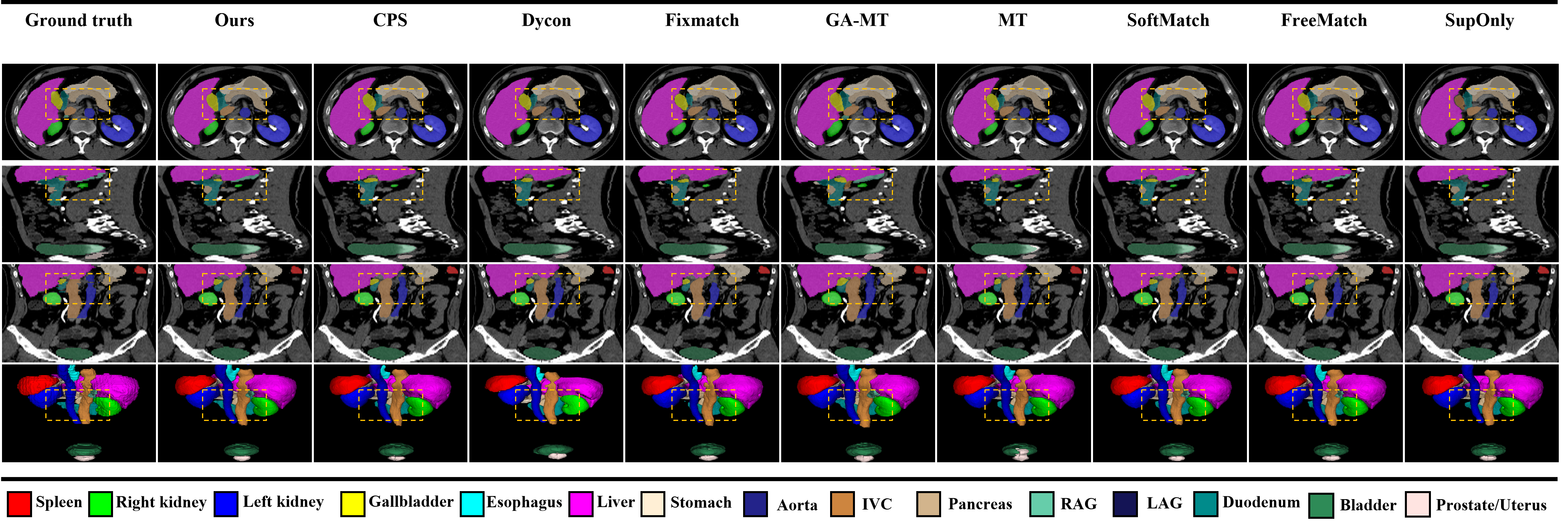}
  \caption{Qualitative comparison of different semi-supervised segmentation
  methods on the AMOS2022 dataset.}
  \label{fig:amos_visual}
\end{figure*}

\textbf{Visual Comparisons.}
Figure \ref{fig:amos_visual} presents a qualitative comparison of segmentation results on the AMOS2022 dataset under 1:5 Ratio setting.
As illustrated across the multi-planar cross-sections (axial, sagittal, and coronal) and the 3D surface renderings, our method generates segmentation maps that align most closely with the Ground Truth.

\subsection{Ablation Studies}
In our ablation study, we validate the core components of ThreshGuide: the labeled proxy, the class-aware threshold, the dynamically adjusted $F_\beta$ components (Base $\beta$ and Error Penalty), and the threshold EMA.
We perform all experiments on the AMOS2022 dataset under the extremely scarce 1:15 labeled ratio setting (80 labeled and 1200 unlabeled scans), with quantitative results summarized in Table~\ref{tab:ablation_components}.

\begin{table}[t]
    \centering
    \footnotesize
    \setlength{\tabcolsep}{2.5pt}
    \renewcommand{\arraystretch}{1.05}

    \begin{tabular}{@{}lccccc cc@{}}
        \toprule
        \multirow{2}{*}{\textbf{ID}}
        & \textbf{Labeled}
        & \textbf{Class-}
        & \textbf{Base}
        & \textbf{Error}
        & \textbf{Thresh.}
        & \multicolumn{2}{c}{\textbf{Metrics (\%)}} \\

        \cmidrule(lr){7-8}

        & \textbf{Proxy}
        & \textbf{aware}
        & \textbf{$\beta$}
        & \textbf{Pen.}
        & \textbf{EMA}
        & \textbf{Dice}
        & \textbf{Jac.} \\

        \midrule

        A1
        & & & & & & 82.63 & 73.99 \\

        A2
        & \checkmark & & & & & 83.40 & 75.29 \\

        A3
        & \checkmark & \checkmark & & & & 83.53 & 75.31 \\

        A4
        & \checkmark & \checkmark & \checkmark & & &
        \sbest{83.69} & \sbest{75.47} \\

        A5
        & \checkmark & \checkmark & \checkmark & \checkmark & &
        83.62 & 75.33 \\

        A6
        & \checkmark & \checkmark & \checkmark & \checkmark & \checkmark &
        \best{84.01} & \best{75.74} \\

        \bottomrule
    \end{tabular}
    \caption{
    Ablation of the core components of ThreshGuide on AMOS2022
    under the 1:15 setting.
    Best and second-best results are shown in \best{red} and \sbest{blue}.
    }
    \label{tab:ablation_components}
\end{table}

\textbf{Effectiveness of the Labeled Proxy.}
The standard semi-supervised \textit{Baseline} (A1) uses a predefined global threshold of 0.95.
This setting cannot adapt to varying learning difficulties or the temporal evolution of the teacher model, yielding a Dice of 82.63\%.
We first introduce the \textit{Global Dynamic} variant (A2), which utilizes the labeled proxy to estimate a single shared threshold online.
This modification improves performance to 83.40\%, demonstrating that labeled data can serve as an effective online reference to provide a reliable signal for pseudo-label selection.

\textbf{Ablation on Core Components.}
We incrementally validate the core components of our proposed ThreshGuide on the AMOS2022 dataset under the extremely scarce 1:15 labeled ratio setting.
``Labeled Proxy'' indicates that threshold targets are dynamically estimated online from teacher predictions on labeled data.
Additionally, ``Base $\beta$'' and ``Error Penalty'' refer to the two key elements of our dynamically adjusted $F_\beta$ optimization (detailed in Sec.~\ref{sec:dynamic_thresholding}).

\textbf{Necessity of Class-Awareness and Base $\beta$.}
Moving beyond a shared threshold, we implement independent thresholds for each class (A3), which slightly raises the mDice to 83.53\%.
We then integrate the foreground-ratio-based Base $\beta$ (A4). This variant achieves 83.69\%, indicating that organs differ substantially in size and precision-coverage preference.
Providing a stable, class-specific prior helps balance the learning pace of multi-organ segmentation.

\textbf{The Coupling of Error Penalty and Temporal Smoothing.}
The critical mechanism in our dynamic $F_\beta$ optimization is the combination of the error penalty and threshold EMA.
Notably, when we directly apply the error penalty without temporal smoothing (A5), the performance slightly drops to 83.62\%.
This phenomenon is highly intuitive: calculating error penalties entirely based on the current batch introduces severe sampling noise.

To resolve this, our full \textit{ThreshGuide} (A6) incorporates Temporal Smoothing (Threshold EMA) to update the class-aware targets before transferring them to unlabeled filtering.
This final step significantly rebounds the performance to the global optimum of 84.01\% Dice and 75.74\% Jaccard.
This validates that the error penalty and threshold EMA are tightly coupled components that must work in synergy to achieve stable and optimal dynamic thresholding.
A component-wise analysis of the inverse-logarithmic base weight and exponential error penalty is provided in Appendix~\ref{sec:appendix_B}.

\begin{table}[t]
    \centering
    \small
    \setlength{\tabcolsep}{16pt}
    \begin{tabular}{@{}lcc@{}}
        \toprule
        \multirow{2}{*}{\textbf{Decay Function}} & \multicolumn{2}{c}{\textbf{Metrics (\%)}} \\
        \cmidrule(l){2-3}
        & \textbf{Dice} & \textbf{Jac.} \\
        \midrule
        Linear             & \sbest{83.19}& \sbest{75.05} \\
        Inverse            & 83.11        & 74.61 \\
        Exponential (Ours) & \best{84.01} & \best{75.74} \\
        \bottomrule
    \end{tabular}
    \caption{
    Ablation study of the decay functions for the error penalty in the dynamic $F_\beta$ optimization on the AMOS2022 dataset (1:15 setting).
    The \best{best} and \sbest{second-best} results are highlighted in \best{red} and \sbest{underlined blue}, respectively.}
    \label{tab:ablation_beta_decay}
\end{table}

\textbf{Ablation on Error Penalty Decay Functions.}
Building on the established synergy between the error penalty and temporal smoothing, we further investigate the specific mathematical formulation of the decay function, as summarized in Table~\ref{tab:ablation_beta_decay}.
Since the error penalty dynamically modulates the weight $\beta$ based on the absolute error of pseudo-labels, the choice of decay function directly impacts the stability of threshold updates.
We compare our default \textit{Exponential} decay against \textit{Linear} and \textit{Inverse} alternatives.

The quantitative results show that both linear and inverse decays yield sub-optimal performance, dropping to 83.19\% and 83.11\% Dice, respectively.
This performance gap can be attributed to the inherent mathematical properties of the exponential function: it provides a smooth, non-linear relaxation that steeply penalizes large initial errors while maintaining a gentle, asymptotic tail.
This specific behavior naturally prevents overly aggressive threshold reductions, ensuring that the error penalty operates with the EMA mechanism to maintain a stable pseudo-label mining process.

\section{Conclusion}
In this paper, we revisit semi-supervised medical image segmentation from the perspective of threshold adaptation and show that labeled data can serve as an effective online proxy for pseudo-label selection.
Built on this proxy, ThreshGuide combines class-aware threshold estimation, dynamic $\beta$ weighting, and threshold EMA to match the class imbalance, reliability variation, and temporal evolution encountered in abdominal multi-organ segmentation.
As confirmed by both the main results and the ablation study, this formulation consistently improves pseudo-label quality and unlabeled data utilization over fixed-threshold heuristics without introducing a hand-crafted threshold schedule.

\bibliographystyle{plainnat}
\bibliography{arXiv}

@article{sohn2020fixmatch,
  title     = {{FixMatch}: Simplifying Semi-Supervised Learning with Consistency and Confidence},
  author    = {Sohn, Kihyuk and Berthelot, David and Carlini, Nicholas and Zhang, Zizhao and Zhang, Han and Raffel, Colin A and Cubuk, Ekin Dogus and Kurakin, Alexey and Li, Chun-Liang},
  journal   = {Advances in Neural Information Processing Systems (NeurIPS)},
  volume    = {33},
  pages     = {596--608},
  year      = {2020}
}

@article{tarvainen2017mean,
  title     = {Mean Teachers are Better Role Models: Weight-Averaged Consistency Targets Improve Semi-Supervised Deep Learning Results},
  author    = {Tarvainen, Antti and Valpola, Harri},
  journal   = {Advances in Neural Information Processing Systems (NeurIPS)},
  volume    = {30},
  year      = {2017}
}

@inproceedings{chen2021semi,
  title     = {Semi-Supervised Semantic Segmentation with Cross Pseudo Supervision},
  author    = {Chen, Xiaokang and Yuan, Yuhui and Zeng, Gang and Wang, Jingdong},
  booktitle = {Proceedings of the IEEE/CVF Conference on Computer Vision and Pattern Recognition (CVPR)},
  pages     = {2613--2622},
  year      = {2021}
}

@inproceedings{jiang2022uncertainty,
  title     = {Uncertainty-Guided Pixel Contrastive Learning for Semi-Supervised Medical Image Segmentation},
  author    = {Jiang, Yicheng and Zhang, Ziqi and He, Ruixuan and Wei, Chenchen and Chen, Zhirui and Pu, Yuge},
  booktitle = {Proceedings of the 31st International Joint Conference on Artificial Intelligence (IJCAI)},
  pages     = {983--990},
  year      = {2022},
  organization = {International Joint Conferences on Artificial Intelligence Organization}
}

@inproceedings{assefa2025dycon,
  title     = {{DyCON}: Dynamic Uncertainty-Aware Consistency and Contrastive Learning for Semi-Supervised Medical Image Segmentation},
  author    = {Assefa, Maregu and Naseer, Muzammal and Ganapathi, Iyyakutti Iyappan and Ali, Syed Sadaf and Seghier, Mohamed L and Werghi, Naoufel},
  booktitle = {Proceedings of the IEEE/CVF Conference on Computer Vision and Pattern Recognition (CVPR)},
  pages     = {30850--30860},
  year      = {2025}
}

@inproceedings{qi2024gradient,
  title={Gradient-aware for class-imbalanced semi-supervised medical image segmentation},
  author={Qi, Wenbo and Wu, Jiafei and Chan, SC},
  booktitle={European Conference on Computer Vision},
  pages={473--490},
  year={2024},
  organization={Springer}
}

@article{ma2024unleashing,
  title={Unleashing the strengths of unlabelled data in deep learning-assisted pan-cancer abdominal organ quantification: the FLARE22 challenge},
  author={Ma, Jun and Zhang, Yao and Gu, Song and Ge, Cheng and Mae, Shihao and Young, Adamo and Zhu, Cheng and Yang, Xin and Meng, Kangkang and Huang, Ziyan and others},
  journal={The Lancet Digital Health},
  volume={6},
  number={11},
  pages={e815--e826},
  year={2024},
  publisher={Elsevier}
}

@inproceedings{cciccek20163d,
  title={3D U-Net: learning dense volumetric segmentation from sparse annotation},
  author={{\c{C}}i{\c{c}}ek, {\"O}zg{\"u}n and Abdulkadir, Ahmed and Lienkamp, Soeren S and Brox, Thomas and Ronneberger, Olaf},
  booktitle={International conference on medical image computing and computer-assisted intervention},
  pages={424--432},
  year={2016},
  organization={Springer}
}

@article{shannon1948,
  title={A mathematical theory of communication},
  author={Shannon, Claude E},
  journal={The Bell system technical journal},
  volume={27},
  number={3},
  pages={379--423},
  year={1948},
  publisher={Nokia Bell Labs}
}

@article{ji2022amos,
  title={Amos: A large-scale abdominal multi-organ benchmark for versatile medical image segmentation},
  author={Ji, Yuanfeng and Bai, Haotian and Ge, Chongjian and Yang, Jie and Zhu, Ye and Zhang, Ruimao and Li, Zhen and Zhanng, Lingyan and Ma, Wanling and Wan, Xiang and others},
  journal={Advances in neural information processing systems},
  volume={35},
  pages={36722--36732},
  year={2022}
}

@article{loshchilov2017decoupled,
  title={Decoupled weight decay regularization},
  author={Loshchilov, Ilya and Hutter, Frank},
  journal={arXiv preprint arXiv:1711.05101},
  year={2017}
}

@article{paszke2019pytorch,
  title={Pytorch: An imperative style, high-performance deep learning library},
  author={Paszke, Adam and Gross, Sam and Massa, Francisco and Lerer, Adam and Bradbury, James and Chanan, Gregory and Killeen, Trevor and Lin, Zeming and Gimelshein, Natalia and Antiga, Luca and others},
  journal={Advances in neural information processing systems},
  volume={32},
  year={2019}
}

@inproceedings{chen2023softmatch,
  title={SoftMatch: Addressing the Quantity-Quality Tradeoff in Semi-supervised Learning},
  author={Chen, Hao and Tao, Ran and Fan, Yue and Wang, Yidong and Wang, Jindong and Schiele, Bernt and Xie, Xing and Raj, Bhiksha and Savvides, Marios},
  booktitle={The Eleventh International Conference on Learning Representations},
  year={2023}
}

@inproceedings{wang2023freematch,
  title={FreeMatch: Self-adaptive Thresholding for Semi-supervised Learning},
  author={Wang, Yidong and Chen, Hao and Heng, Qiang and Hou, Wenxin and Fan, Yue and Wu, Zhen and Wang, Jindong and Savvides, Marios and Shinozaki, Takahiro and Raj, Bhiksha and others},
  booktitle={Eleventh International Conference on Learning Representations},
  year={2023},
  organization={OpenReview. net}
}

@inproceedings{he2023geometric,
  title={Geometric visual similarity learning in 3d medical image self-supervised pre-training},
  author={He, Yuting and Yang, Guanyu and Ge, Rongjun and Chen, Yang and Coatrieux, Jean-Louis and Wang, Boyu and Li, Shuo},
  booktitle={Proceedings of the IEEE/CVF Conference on Computer Vision and Pattern Recognition},
  pages={9538--9547},
  year={2023}
}

\clearpage
\appendix

\section{Implementation and Architecture Details}
\label{sec:appendix_A}

To ensure a reliable evaluation and reproducibility, we report the average results across three random seeds (0, 2025, and 2026).

\subsection{Algorithm Pseudocode}
\label{sec:A_pseudocode}

Algorithm~\ref{alg:threshguide} summarizes the training procedure of ThreshGuide, using notation consistent with the main paper (Section~\ref{sec:method}).

\begin{algorithm*}[t]
\caption{ThreshGuide Training Procedure}
\label{alg:threshguide}
\begin{algorithmic}[1]
\REQUIRE Labeled set $\mathcal{D}^{l}$, unlabeled set $\mathcal{D}^{u}$, number of classes $C$, unsupervised weight $\lambda_u$, threshold momentum $\alpha_{\tau}$, teacher momentum cap $\alpha_{\max}$
\ENSURE Trained student $f_{\theta}$ and teacher $f_{\phi}$
\STATE Initialize student $f_{\theta}$ (Kaiming normal), teacher $f_{\phi} \leftarrow f_{\theta}$, thresholds $\tau_c^{(0)} \leftarrow \tau_0$, and foreground ratios $\mu_c^{(0)} \leftarrow \frac{1}{C-1}$
\FOR{each training iteration $t$}
    \STATE Sample a mini-batch of labeled pairs $(x^l, y^l)$ and unlabeled volumes $x^u$
    \STATE \textbf{Supervised branch:} compute $\mathcal{L}_{s}$ on the labeled batch (Eq.~\eqref{eq:sup_loss})
    \STATE \textbf{Labeled-proxy calibration:} teacher forward on $x^l$ without gradient; update $\mu_c^{(t)}$ and the normalized error $\delta_c^{(t)}$, then $\beta_c^{(t)}$ (Eqs.~\eqref{eq:class_occupancy}--\eqref{eq:beta_final})
    \STATE For each class, sort the confidences of its predicted voxels and set the proxy target $\hat{\tau}_c^{(t)}$ to the top-$k$ cutoff that maximizes $F_{\beta_c}$ (Eqs.~\eqref{eq:pre_cov}--\eqref{eq:proxy_target})
    \STATE Update thresholds $\tau_c^{(t)} \leftarrow \alpha_{\tau}\tau_c^{(t-1)} + (1-\alpha_{\tau})\hat{\tau}_c^{(t)}$ for the classes present in the batch (Eq.~\eqref{eq:threshold_ema})
    \STATE \textbf{Unsupervised branch:} teacher forward on the weak view of $x^u$; accept a voxel if its maximum confidence reaches the threshold of its predicted class, and supervise the student prediction on the strong view at accepted voxels (Eqs.~\eqref{eq:class_mask}--\eqref{eq:unsup_loss})
    \STATE Update $\theta$ with $\mathcal{L}_{total} = \mathcal{L}_{s} + \lambda_u\mathcal{L}_{u}$ (Eq.~\eqref{eq:total_loss}); update the teacher by EMA with $\alpha_t = \min(1 - \frac{1}{t+1},\, \alpha_{\max})$ (Eq.~\eqref{eq:teacher_ema})
\ENDFOR
\end{algorithmic}
\end{algorithm*}

\subsection{Datasets}
\label{sec:A_datasets}

We evaluate ThreshGuide on two publicly available abdominal multi-organ CT benchmarks.

\textbf{FLARE2022}~\citep{ma2024unleashing}\footnote{\url{https://flare22.grand-challenge.org/}}
comprises 100 labeled and 2{,}000 unlabeled CT scans covering 13 organ classes (with one background): the liver, spleen, pancreas, right kidney, left kidney, stomach, gallbladder, esophagus, aorta, inferior vena cava, right adrenal gland, left adrenal gland, and duodenum.
The labeled scans are split into 60, 20, and 20 for training, validation, and testing, respectively.

\textbf{AMOS2022}~\citep{ji2022amos}\footnote{\url{https://amos22.grand-challenge.org/}}
is a 16-class segmentation dataset targeting 15 anatomical structures, including two additional organs not found in FLARE2022: the bladder and prostate/uterus.
Its 300 labeled scans are partitioned into 240, 30, and 30 for training, validation, and testing, respectively.
In addition, the dataset includes 1{,}200 unlabeled scans.

\subsection{Data Preprocessing}
\label{sec:A_preprocessing}

All raw volumes are preprocessed identically for both datasets with a unified pipeline consisting of three steps applied to each volume:
\begin{enumerate}
    \item \textbf{Orientation standardization.}
    Each volume is reoriented to the canonical RAS (Right--Anterior--Superior) axis code, ensuring a consistent anatomical coordinate system across all subjects and scanners.

    \item \textbf{Intensity windowing and normalization.}
    Voxel intensities are clipped to $[-40, 325]$~HU to suppress out-of-range artifacts (e.g., metal implants or air).
    A per-volume z-score normalization (zero mean, unit variance) is then applied, followed by min-max rescaling to $[0, 1]$.

    \item \textbf{Spatial resampling.}
    Volumes are resampled to a uniform target spacing of $(1.2548, 1.2548, 2.50)$~mm.
    Images use cubic interpolation to preserve smooth intensity transitions; annotation masks use nearest-neighbor interpolation to retain integer label values.
\end{enumerate}
The preprocessed volumes are stored in HDF5 format with chunked, gzip-compressed access, so that a single $64\times128\times128$ crop can be read without decompressing the entire volume; this reduces per-iteration I/O latency by roughly $3\times$ compared with loading full volumes at training time.

\subsection{Optimization Settings}
\label{sec:A_optimization}

We optimize the segmentation network using AdamW~\citep{loshchilov2017decoupled} with momentum parameters $\beta=(0.9, 0.999)$ and a weight decay of $10^{-3}$.
The learning rate follows a polynomial schedule $(1 - \frac{t}{T})^{0.9}$, where $t$ and $T$ denote the current and total iterations, respectively.
The initial learning rate is set to $1 \times 10^{-1}$.
For the overall training objective, the unsupervised loss weight is set to $\lambda_u=0.1$.
The models are trained for a total of 37{,}500 and 45{,}000 iterations on FLARE2022 and AMOS2022, respectively.

\subsection{Computational Cost}
\label{sec:A_cost}

We evaluate the computational footprint of ThreshGuide on a single NVIDIA RTX 4090 GPU (24\,GB VRAM) under the AMOS2022 1:15 setting, using a batch size of 8 (4 labeled and 4 unlabeled $64\times128\times128$ crops).
The workstation is equipped with an AMD EPYC 7542 32-core CPU and 504\,GB system RAM, and runs PyTorch 2.10.0.
The steady-state per-iteration wall-clock time is approximately \textbf{0.40\,s}.
Within this budget, the teacher's extra forward pass on labeled crops and the per-class $F_\beta$ rank-search consume roughly 15\%.
When accounting for periodic overheads such as evaluation and checkpointing (3--4 s per few hundred iterations), the amortized cost amounts to roughly 1.0 s per iteration.
Under this throughput, training for 45,000 iterations requires 13.2 hours in total, compared to 13.0 hours for the FixMatch baseline.
This constitutes a marginal relative computational overhead of less than 2\%.

Peak VRAM usage is \textbf{12.43\,GB}, well within the 24\,GB budget of a single consumer GPU.
The steady-state allocated memory is 4.39\,GB, with PyTorch's CUDA caching allocator reserving 14.76\,GB.
This modest footprint leaves headroom for larger input patch sizes or deeper backbones without resorting to gradient checkpointing or model parallelism.

\subsection{Evaluation Metrics}
\label{sec:A_metrics}

We adopt two standard volumetric overlap metrics to quantify segmentation accuracy.
Let $\hat{Y}_c$ and $Y_c$ denote the predicted and ground-truth binary masks for class $c$, respectively.
The \textbf{Dice Similarity Coefficient} (DSC) is defined as
\begin{equation}
\mathrm{Dice}_c = \frac{2\,|\hat{Y}_c \cap Y_c|}{|\hat{Y}_c| + |Y_c|},
\end{equation}
and the \textbf{Jaccard Index} (IoU) as
\begin{equation}
\mathrm{Jac}_c = \frac{|\hat{Y}_c \cap Y_c|}{|\hat{Y}_c \cup Y_c|}.
\end{equation}
Both metrics range from $0$ (no overlap) to $1$ (perfect agreement).
We report the class-wise Dice for each organ and the overall average Dice and Jaccard across all foreground classes.
The Dice coefficient is chosen as the primary metric because it is less sensitive to the extreme foreground--background imbalance inherent in small abdominal organs, and it is the standard benchmark in prior abdominal segmentation literature, enabling direct comparison with existing methods.
The Jaccard index is reported alongside Dice as a complementary measure: because $\mathrm{Jac}_c = \mathrm{Dice}_c / (2 - \mathrm{Dice}_c)$, it penalizes false positives and false negatives more stringently than Dice, providing a stricter assessment of boundary quality for small organs where even minor over-segmentation noticeably degrades IoU.

\section{Theoretical Analysis}
\label{sec:appendix_B}

Let $P$ denote Precision and $R$ denote Coverage.
The $F_\beta$ score used in the main text is
\begin{equation}\label{eq:A_fbeta}
F_\beta(P, R) = \frac{(1 + \beta^2) P R}{\beta^2 P + R}
= \frac{1 + \beta^2}{\dfrac{\beta^2}{R} + \dfrac{1}{P}},
\end{equation}
which is a \textbf{weighted harmonic mean}: $1/R$ receives weight $\beta^2$, $1/P$ receives weight $1$.

We choose the harmonic mean over the arithmetic mean $(P+R)/2$ because of its behavior at extreme values.
If the model selects an extremely high threshold, $P \approx 1.0$ but $R \ll 0.01$; the arithmetic mean still yields $\approx 0.505$, masking the collapse of unlabeled supervision.
By contrast, the harmonic mean satisfies
\begin{equation}
\lim_{R \to 0} F_\beta = 0, \qquad \lim_{P \to 0} F_\beta = 0.
\end{equation}
This prevents degenerate solutions in which a single metric is maximized at the expense of the other.
If either precision or coverage collapses, $F_\beta$ approaches zero, so the threshold selected by maximizing $F_\beta$ must keep both away from zero.

\subsection{Dynamic $\beta_c$ and Phase-Dependent Behavior}
\label{sec:A_beta_behavior}

Recall from the main text that the class-wise weight is
\begin{equation}\label{eq:A_beta}
\beta_c^{(t)} = \underbrace{\frac{1}{1 - \log(\mu_c^{(t)})}}_{\beta_{\text{base}}^{(c)}} \cdot \underbrace{\exp\!\bigl(-\delta_c^{(t)}\bigr)}_{\text{error penalty}},
\end{equation}
where $\mu_c^{(t)}$ is the EMA-smoothed foreground proportion and $\delta_c^{(t)}$ is the normalized teacher error.

From Eq.~\eqref{eq:A_fbeta}, the sensitivity of $F_\beta$ to $P$ and $R$ is governed by:
\begin{equation}
\frac{\partial F_\beta}{\partial (1/P)} \propto 1,\qquad \frac{\partial F_\beta}{\partial (1/R)} \propto \beta^2.
\end{equation}
This induces two regimes:
\begin{itemize}
    \item \textbf{Early training (conservative).}
    For small organs, $\mu_c^{(t)}$ is tiny and $\delta_c^{(t)}$ is large, so $\beta_c \ll 1$.
    The denominator is dominated by $1/P$, enforcing a strict high threshold that suppresses false positives when the teacher is unreliable.

    \item \textbf{Late training (explorative).}
    As $\delta_c^{(t)} \to 0$ and $\mu_c^{(t)}$ stabilizes, $\beta_c$ relaxes toward $1$.
    The weight shifts toward $1/R$, lowering the threshold to mine harder unlabeled samples safely.
\end{itemize}

\subsection{A Dual-Level Minimax/Maximin Interpretation}
\label{sec:A_minimax}

We give a heuristic interpretation of the ThreshGuide objective at two levels: per class and across classes.

\medskip
\noindent
\textbf{Level 1: The Per-Class Precision-Coverage Trade-Off.}

Threshold selection for a given class involves a trade-off between two competing criteria:
\begin{itemize}
    \item \textbf{Minimizing error:} Reducing the incidence of incorrect pseudo-labels, equivalent to maximizing precision ($P$).
    \item \textbf{Maximizing coverage:} Increasing the proportion of ground-truth voxels utilized for supervision, equivalent to maximizing coverage ($R$).
\end{itemize}

These objectives conflict: elevated thresholds improve precision at the cost of data utilization, while lowered thresholds increase coverage but introduce labeling noise.

A standard way to balance these goals is the symmetric maximin formulation $\max_\tau \min\bigl(P(\tau), R(\tau)\bigr)$.
Because the $\min$ function is non-smooth, the $F_\beta$ score serves as a differentiable surrogate.
For $P, R > 0$, $\min(P, R) \leq F_\beta \leq \max(P, R)$, and $F_\beta \rightarrow 0$ if either metric collapses.

Here $\beta$ acts as an adaptive weight: maximizing $F_\beta$ approximates an \textbf{asymmetric soft maximin objective}, $\max_\tau \min\bigl(\beta^2 P(\tau), R(\tau)\bigr)$.
Adjusting $\beta$ shifts the balance between error suppression and coverage expansion over the course of training.

\medskip
\noindent
\textbf{Level 2: Cross-Class Risk as a Minimax Objective.}

At the dataset level, balancing multi-class performance resembles a minimax risk formulation:
$\min_{\boldsymbol{\tau}} \max_{c \in \{1,\ldots,C\}}\, \mathcal{R}_c(\tau_c)$,
where $\boldsymbol{\tau} = \{\tau_1, \ldots, \tau_C\}$ and $\mathcal{R}_c(\tau_c)$ represents the empirical pseudo-label risk for class $c$.

The per-class weight $\beta_c$ provides a heuristic relaxation of this global objective.
For high-risk classes---typically minority classes with small foreground ratios $\mu_c$ and large error penalties $\delta_c$---$\beta_c$ is strongly suppressed, which skews the per-class objective toward precision, yields a tighter threshold $\tau_c$, and alleviates the worst-case class errors.
In contrast, lower-risk classes receive a larger $\beta_c$ and broader coverage.

Note that $\beta_c$ is updated independently for each class and involves no cross-class gradient computation, so this formulation is a heuristic relaxation of the cross-class minimax problem rather than an exact solver.

\subsection{Mathematical Properties of the $\beta_c$ Components}
\label{sec:A_beta_formulation}

\noindent
\textbf{Inverse-Logarithmic Transform.}

Organ volumes naturally exhibit a long-tailed distribution.
A direct linear mapping of the foreground ratio $\mu_c^{(t)}$ would disproportionately compress small-organ weights toward zero, inducing numerical instability.
The transformation $\beta_{\text{base}}^{(c)} = 1 / (1 - \log(\mu_c^{(t)}))$ addresses this by logarithmically calibrating the dynamic range. Specifically, as $\mu_c \to 1$, $\beta_{\text{base}}$ approaches $1$, smoothly transitioning to a balanced $F_1$ objective.
Conversely, as $\mu_c \to 0$, $\beta_{\text{base}}$ exhibits a gradual logarithmic decay toward $0$.
This gradual decay avoids an abrupt collapse of the weight while still shifting the objective toward precision for tiny anatomical structures.

\noindent
\textbf{Exponential Error Penalty.}

The exponential formulation $\exp(-\delta_c^{(t)})$ remains in $(0,1]$ and decays monotonically as $\delta_c^{(t)}$ increases.
When the teacher is reliable ($\delta_c \to 0$), the penalty approaches $1$ and barely modifies $\beta_c$.
When the teacher produces severe errors ($\delta_c \gg 1$), the penalty decays rapidly toward $0$: $\beta_c$ becomes small, the $F_\beta$ objective then weights precision more heavily, and the selected threshold $\tau_c$ rises, so few noisy pseudo-labels are accepted.
In contrast, a linear penalty ($1 - \delta_c$) can become negative when $\delta_c > 1$ and does not distinguish moderate from large errors as sharply as the exponential form.

\section{Evolution of Class-Aware Thresholds}
\label{sec:appendix_C}

\begin{figure*}[t]
    \centering
    \includegraphics[width=0.96\textwidth]{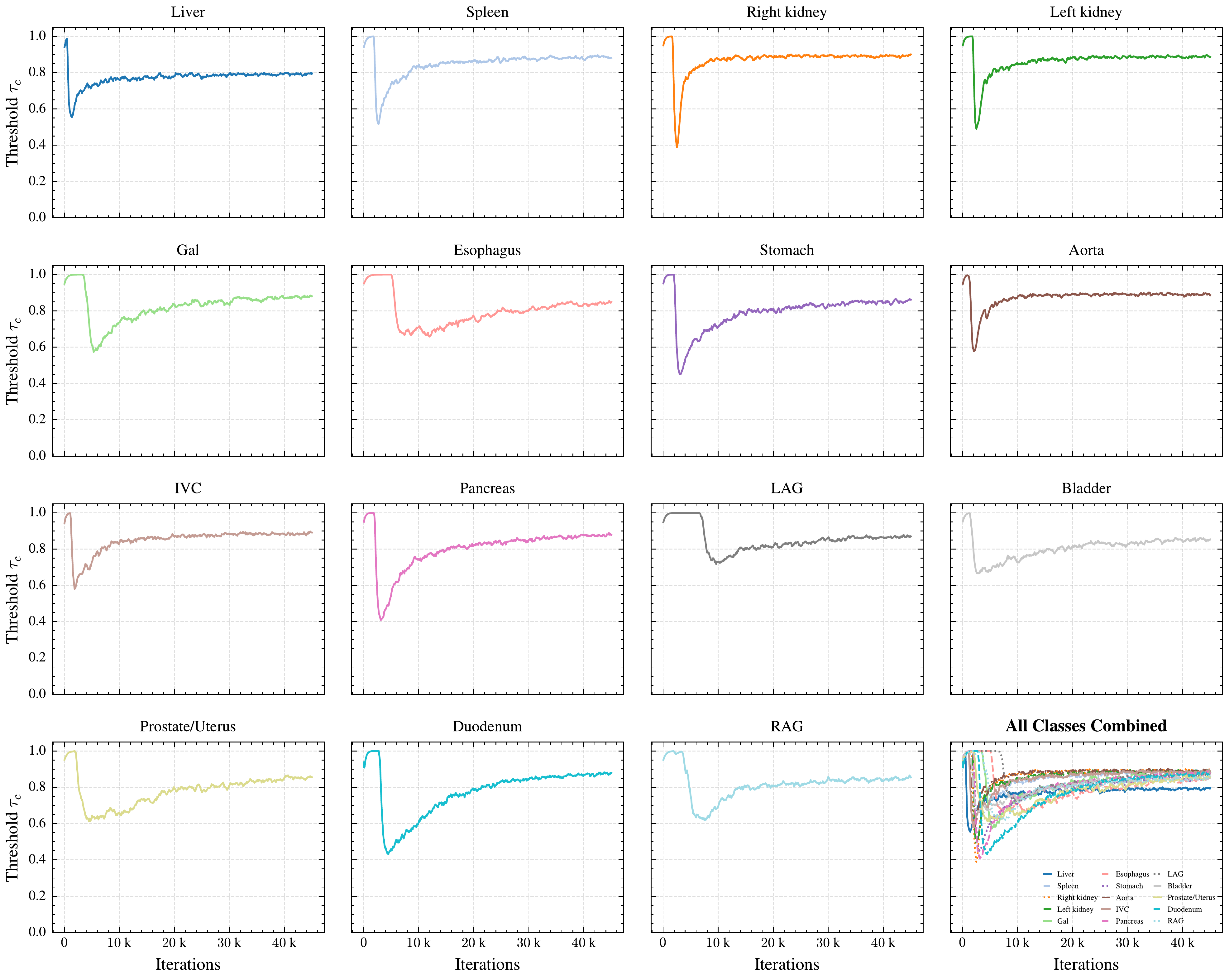}
    \caption{Evolution of the class-aware confidence thresholds $\tau_c$
    throughout the training process.
    The grid displays individual threshold trajectories for all 15 foreground
    organs, alongside a combined overall view in the bottom right.
    Most classes follow a ``high-drop-recovery'' pattern: harder organs
    (e.g., Duodenum and Stomach) show much deeper threshold drops,
    while all classes recover to higher values in the later stages.}
    \label{fig:threshold_evolution}
\end{figure*}

We visualize the evolution of the class-aware thresholds $\tau_c$ during training to illustrate the phase-dependent behavior described in Appendix~\ref{sec:A_beta_behavior}.

Figure~\ref{fig:threshold_evolution} shows the threshold trajectories for all 15 foreground organs, together with a combined overall view.
For most classes the trajectory follows the same ``high-drop-recovery'' shape, but the depth and timing of the drop differ across classes, following the class-dependent $\beta_c$ in Eq.~\eqref{eq:A_beta}.

\textbf{Phase 1: Conservative Initialization.}
During the early stages of training, the thresholds remain at a stringent level (approaching $1.0$).
The teacher is initially unreliable, so the large prediction error $\delta_c^{(t)}$ drives the exponential error penalty $\exp(-\delta_c^{(t)})$ toward zero.
This makes $\beta_c \to 0$ and the $F_\beta$ objective precision-oriented.
As a result, few pseudo-labels are accepted at this stage, which limits confirmation bias while the learned representations are still fragile.

\textbf{Phase 2: Class-Specific Exploration.}
As training progresses and basic feature representations are established, the decreasing error $\delta_c^{(t)}$ weakens the error penalty and the thresholds can drop.
The depth of the drop depends on the class-specific base weight $\beta_{\text{base}}^{(c)} = 1 / (1 - \log(\mu_c^{(t)}))$.
For large or easy-to-learn organs (e.g., Liver), the large foreground ratio $\mu_c$ drives $\beta_{\text{base}}^{(c)}$ toward $1$, which yields a balanced precision-coverage trade-off and a shallower drop ($\tau_c \approx 0.6$).
For smaller or more difficult targets (e.g., Duodenum, Stomach, and Pancreas), $\mu_c$ is much smaller, and maintaining coverage on these sparse classes requires a deeper drop (to roughly $0.45$--$0.50$).
This relaxation admits more pseudo-labels for these classes and improves their coverage.

\textbf{Phase 3: Mature Refinement.}
In the later stages of training, $\delta_c^{(t)} \to 0$ and the class proportion $\mu_c^{(t)}$ stabilizes.
With both statistics at a plateau, the threshold targets again favor high-confidence regions: the thresholds recover to higher values (between $0.80$ and $0.90$ for most classes), filtering out the remaining noise.

This evolving schedule suggests that a fixed threshold such as the global $\tau=0.95$ used by FixMatch is a poor fit for this task, and that the precision--coverage balance needs to be set per class and adjusted over the course of training.

\begin{figure*}[t]
    \centering
    \includegraphics[width=0.98\textwidth]{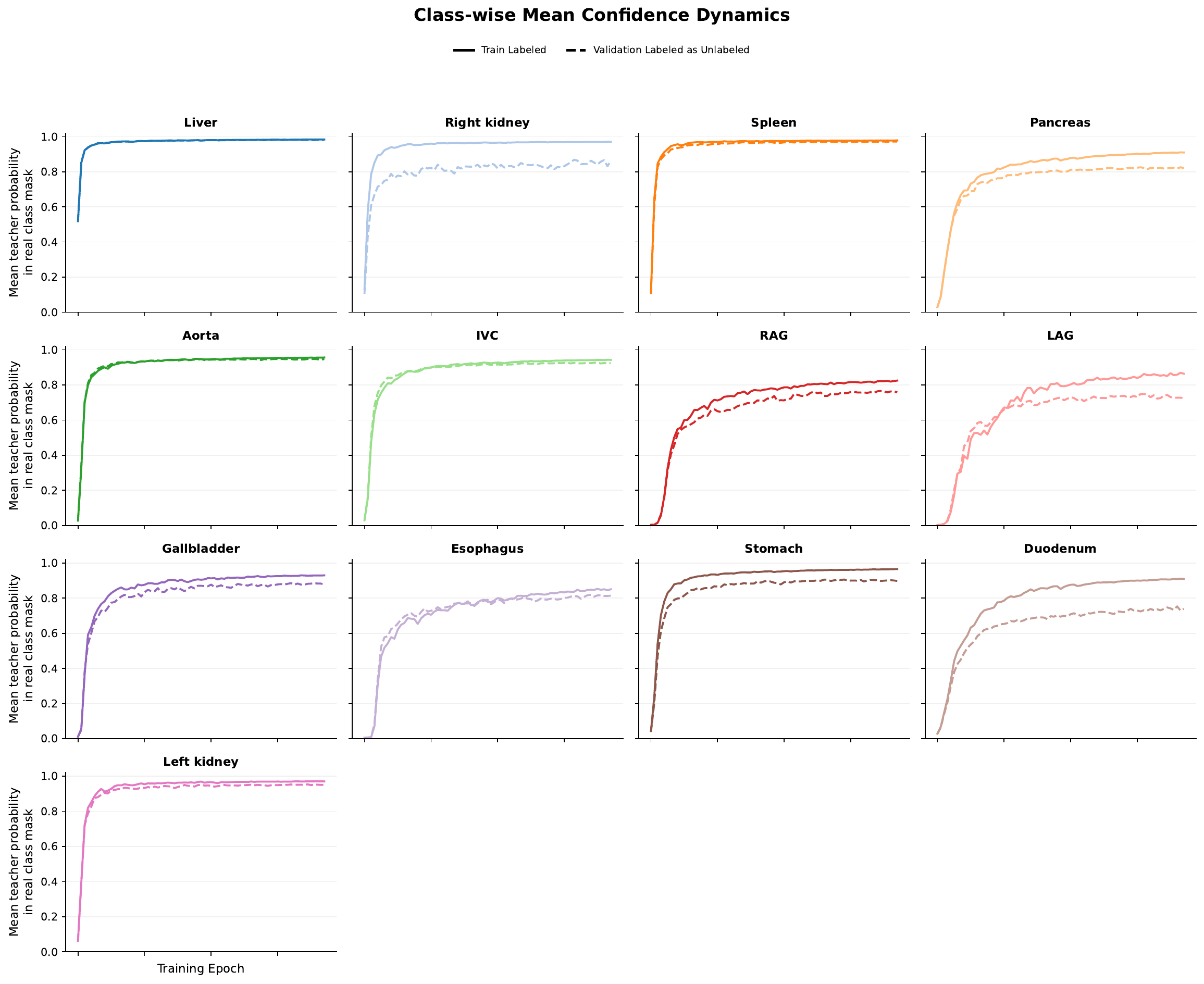}
    \caption{Class-wise mean true-class confidence dynamics on randomly
    sampled patches from AMOS2022.
    Each panel corresponds to one foreground organ.
    Solid curves denote the teacher's mean predicted probability for the
    ground-truth class within labeled training patches; dashed curves
    denote the same metric evaluated on validation patches treated as
    proxy-unlabeled data.
    A persistent cross-split confidence gap is observable for every class:
    the teacher assigns substantially higher confidence to labeled samples
    than to unlabeled ones throughout training.
    Moreover, the magnitude and convergence rate of this gap vary
    considerably across organs---large structures (e.g., Liver, Spleen)
    exhibit a narrow gap that closes early, whereas small or
    morphologically variable organs (e.g., Duodenum, Stomach, Pancreas)
    retain a wide gap even in later epochs.
    This class-dependent behavior motivates the use of
    independent per-class thresholds rather than a single global value.}
    \label{fig:confidence_trends}
\end{figure*}

\textbf{Cross-Split Confidence Analysis.}
We further track the evolution of the class-wise mean true-class confidence---defined as the teacher's predicted probability for the ground-truth class, averaged over all voxels within the corresponding annotation mask---on both the labeled training set and the validation set (a proxy for unlabeled data) throughout training.
Figure~\ref{fig:confidence_trends} presents the per-class confidence trajectories for all 15 foreground organs, with each panel displaying the labeled (solid) and pseudo-unlabeled (dashed) curves for a single anatomical class.

First, a consistent cross-split gap exists for every class: the teacher assigns higher true-class confidence to labeled samples than to unlabeled ones at all training stages, so a threshold calibrated solely on labeled data would over-filter unlabeled predictions.
Second, the gap is class-dependent: large, high-contrast organs such as Liver and Spleen exhibit a narrow gap that diminishes within the first few epochs, whereas small or morphologically variable structures (e.g., Duodenum, Stomach, Pancreas, and the adrenal glands) maintain a wide gap throughout training.
Third, the rate at which the gap closes is non-uniform: some classes stabilize early while others continue to change well into the later training phases.

These observations suggest that a single global threshold cannot accommodate the different learning paces of the anatomical classes; our method addresses this by adapting $\tau_c$ independently to each organ's evolving confidence distribution (Eq.~\eqref{eq:A_beta}).

\begin{figure*}[t]
    \centering
    \includegraphics[width=0.98\textwidth]{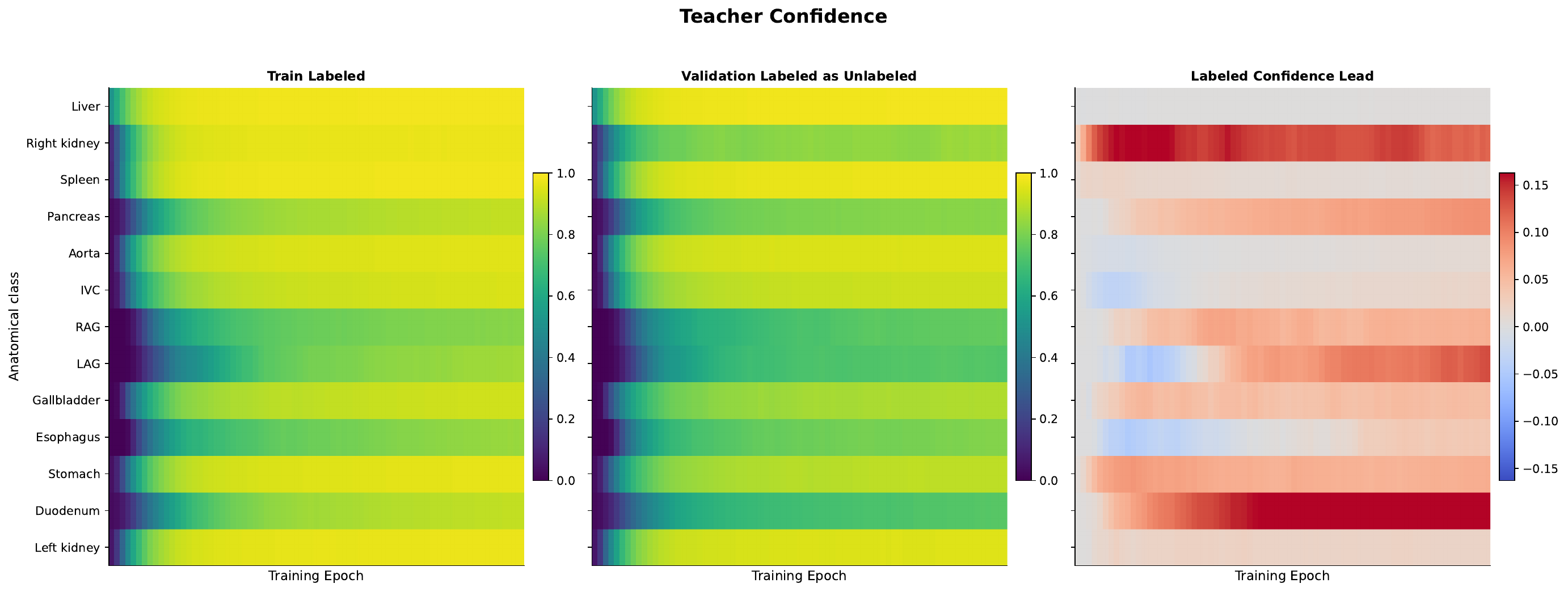}
    \caption{Class-by-epoch confidence heatmaps of the teacher model on
    AMOS2022.
    \textbf{Left:} EMA-smoothed mean true-class confidence on labeled
    training patches (viridis scale, $[0, 1]$).
    \textbf{Middle:} the same metric on validation patches treated as
    unlabeled data.
    \textbf{Right:} the confidence lead (train minus validation),
    visualized with a diverging colormap centered at zero; warm colors
    indicate classes where the teacher is substantially more confident on
    labeled data.
    Rows correspond to the 15 foreground organs and columns to training
    epochs.
    The persistent positive lead for difficult classes (bottom rows)
    indicates that a global threshold would over-filter their unlabeled
    predictions throughout training.}
    \label{fig:confidence_heatmaps}
\end{figure*}

\textbf{Class-by-Epoch Confidence Landscape.}
Figure~\ref{fig:confidence_heatmaps} arranges the same statistics as a class-by-epoch matrix, which makes cross-class comparison easier than the per-class panels of Figure~\ref{fig:confidence_trends}.

The left panel shows a hierarchy across classes: rows for large organs (Liver, Spleen, Kidneys) approach confidence $1$ within the first few epochs, whereas rows for small or ambiguous structures (Pancreas, Duodenum, Adrenal Glands) remain much lower throughout training. The middle panel shows the same ordering on the validation split at uniformly lower levels, reflecting the gap between seen and unseen data.

The right panel shows the confidence lead (train minus validation) with a diverging colormap. The lead is positive for all classes at all epochs and highly non-uniform: it is small and decays quickly for easy classes, but remains large for difficult classes with little sign of closing by the end of training. A single scalar threshold would thus be too permissive for well-learned classes and too restrictive for difficult ones, which is what the per-class thresholds in ThreshGuide are designed to address.

\section{Additional Qualitative Results on FLARE2022}
\label{sec:appendix_D}

\begin{figure*}[t]
    \centering
    \includegraphics[width=0.98\textwidth]{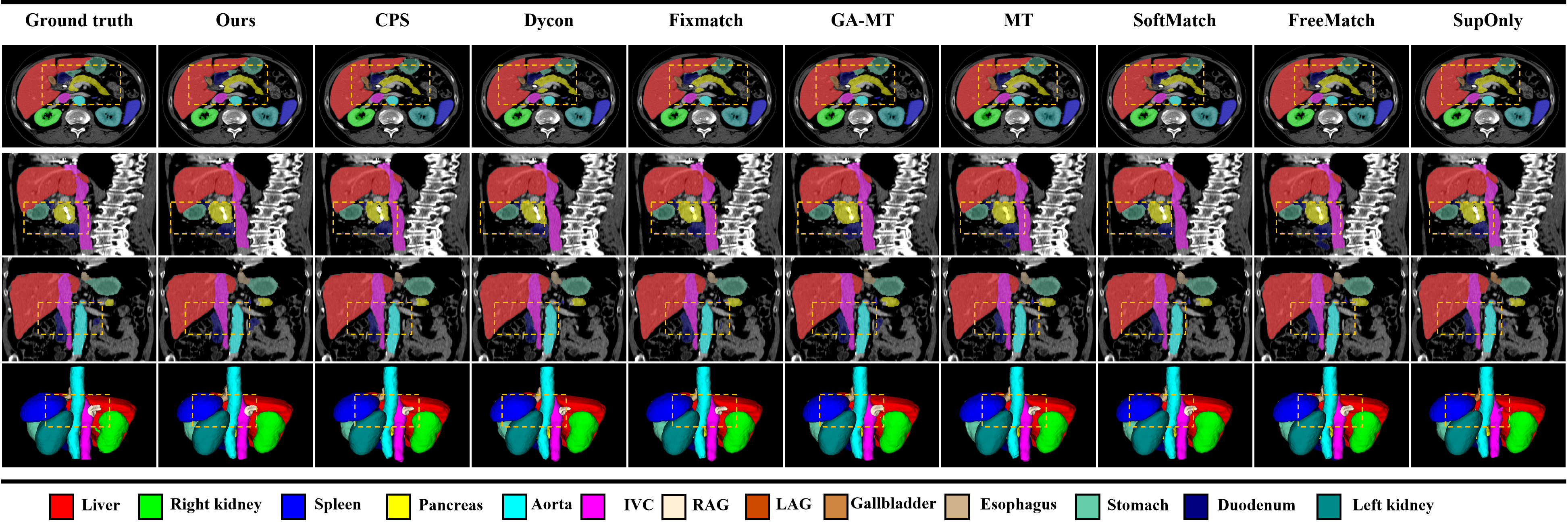}
    \caption{Qualitative comparison of different semi-supervised
    segmentation methods on the FLARE2022 dataset under the 60:2000
    (1:33) labeled ratio (seed 2026).
    Multi-planar cross-sections (axial, sagittal, and coronal) and 3D
    surface renderings are shown for a representative case.
    ThreshGuide produces segmentation maps that align most closely with
    the ground truth, particularly for small and challenging structures
    (e.g., Gallbladder, Esophagus, and Duodenum) where competing methods
    exhibit noticeable under-segmentation or boundary irregularities.}
    \label{fig:flare_visual}
\end{figure*}

To complement the quantitative results reported in the main paper (Table~\ref{tab:main_results_flare}), we present a qualitative comparison on the FLARE2022 dataset in Figure~\ref{fig:flare_visual}.
The visualization is generated from the model trained with seed 2026 under the 60:2000 (1:33) labeled ratio.

Consistent with the observations on AMOS2022, ThreshGuide yields segmentation maps that most closely resemble the ground truth across all three orthogonal views and the 3D surface rendering.
The improvement is particularly pronounced for small, low-contrast organs such as the Gallbladder, Esophagus, and Duodenum, where competing methods (e.g., FixMatch, CPS, and DyCON) exhibit visible under-segmentation, fragmented predictions, or irregular boundaries.
In contrast, ThreshGuide produces compact masks with smoother contours, consistent with the lower thresholds it assigns to these difficult organs.

\end{document}